\documentclass[10pt,twocolumn,letterpaper]{article}

\usepackage{iccv}              
\usepackage{bm}
\definecolor{iccvblue}{rgb}{0.21,0.49,0.74}
\usepackage[pagebackref,breaklinks,colorlinks,allcolors=iccvblue]{hyperref}
\usepackage{endnotes}
\usepackage{multirow}
\usepackage{algorithm}
\usepackage{algpseudocode}
\def\paperID{4832} 
\def\confName{ICCV}
\def\confYear{2025}

\title{Weakly Supervised Visible-Infrared Person Re-Identification via Heterogeneous Expert Collaborative Consistency Learning}

\author{%
  Yafei Zhang\textsuperscript{1}, %
  Lingqi Kong\textsuperscript{1}, %
  Huafeng Li\textsuperscript{1}\thanks{Corresponding author: Huafeng Li (hfchina99@163.com).}, %
  Jie Wen\textsuperscript{2}%
  \\
  \textsuperscript{1}Faculty of Information Engineering and Automation, Kunming University of Science and Technology\\
  \textsuperscript{2}School of Computer Science and Technology, Harbin Institute of Technology, Shenzhen\\
  {\tt\small zyfeimail@163.com, lingqikong2333@gmail.com, hfchina99@163.com, jiewen\_pr@126.com}}

\begin{document}
\maketitle
\begin{abstract}
To reduce the reliance of visible-infrared person re-identification (ReID) models on labeled cross-modal samples, this paper explores a weakly supervised cross-modal person ReID method that uses only single-modal sample identity labels, addressing scenarios where cross-modal identity labels are unavailable. To mitigate the impact of missing cross-modal labels on model performance, we propose a heterogeneous expert collaborative consistency learning framework, designed to establish robust cross-modal identity correspondences in a weakly supervised manner. This framework leverages labeled data from each modality to independently train dedicated classification experts. To associate cross-modal samples, these classification experts act as heterogeneous predictors, predicting the identities of samples from the other modality. To improve prediction accuracy, we design a cross-modal relationship fusion mechanism that effectively integrates predictions from different experts. Under the implicit supervision provided by cross-modal identity correspondences, collaborative and consistent learning among the experts is encouraged, significantly enhancing the model's ability to extract modality-invariant features and improve cross-modal identity recognition. Experimental results on two challenging datasets validate the effectiveness of the proposed method. Code is available at \url{https://github.com/KongLingqi2333/WSL-VIReID}.
\end{abstract}
\vspace{-3.5mm}
\section{Introduction}
\label{sec:intro}
Visible-Infrared Person Re-Identification (VIReID), a core task in cross-modal person re-identification (ReID), plays a pivotal role in intelligent security surveillance and smart city construction. This task involves matching pedestrian images from one modality (infrared or visible) with those from the other to accurately identify target pedestrians. It effectively overcomes the limitations of RGB-based ReID in nighttime and low-light conditions. However, due to the inherent differences in imaging principles between visible and infrared images, significant visual discrepancies exist between them. These modality-specific differences not only make annotating cross-modal datasets significantly more challenging and costly but also pose formidable challenges for cross-modal pedestrian retrieval.
\begin{figure}[t!]
	\centering
	\includegraphics[height=1.8in]{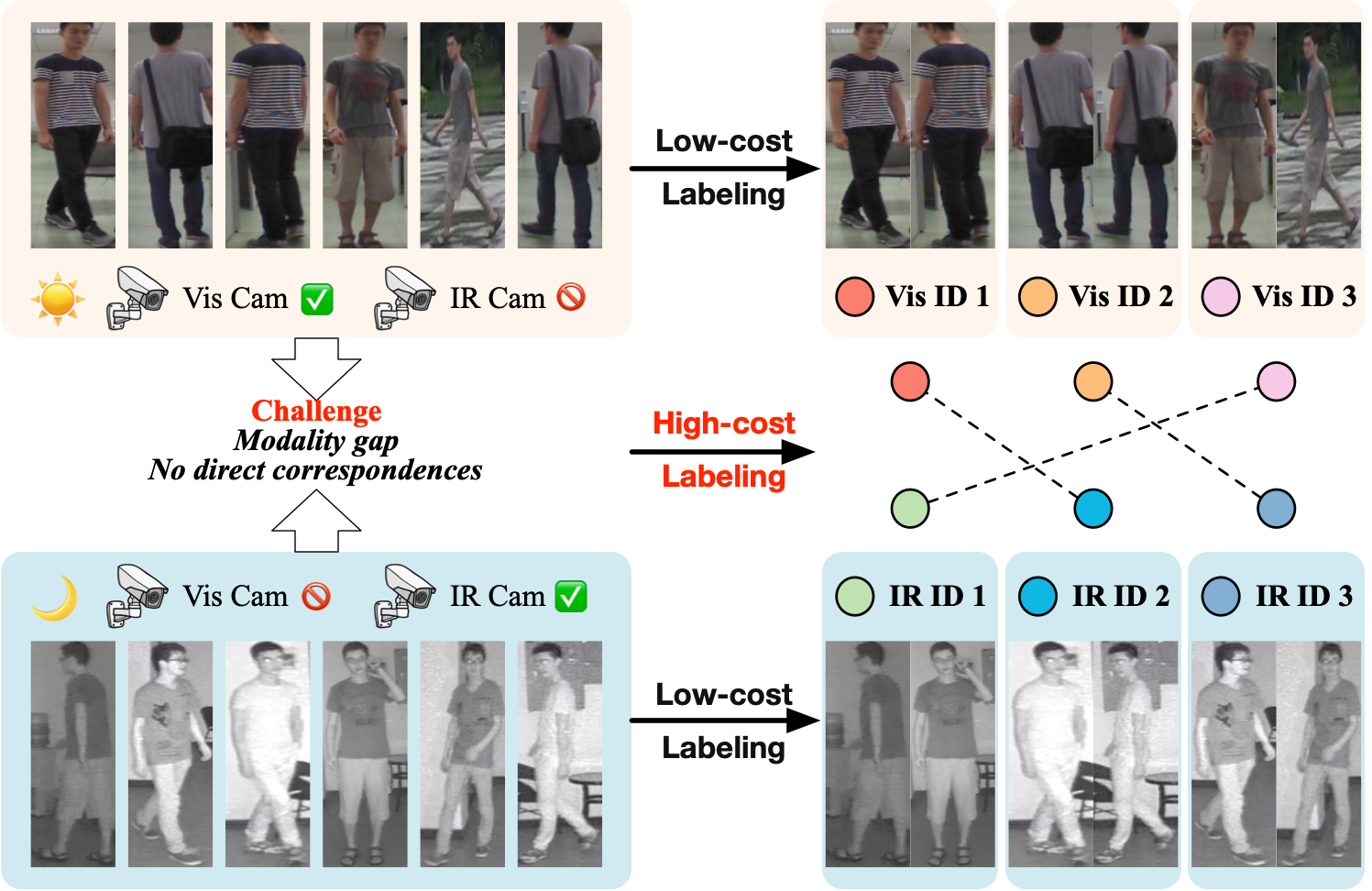}\vspace{-1.5mm}
	\caption{VIReID faces two key challenges: asynchronous visible-infrared (VIS-IR) camera operation causes missing cross-modal identity correspondences, while large appearance differences raise annotation costs beyond those in single-modality scenarios.}\vspace{-6.5mm}
	\label{fig1}
\end{figure}

Existing approaches typically fall into three learning paradigms: fully supervised, semi-supervised, and unsupervised. Fully supervised approaches \cite{1,2,3,4} rely on accurately labeled cross-modal identity samples for model training. Semi-supervised approaches \cite{5,6,7} leverage a combination of partially labeled and unlabeled data, thereby reducing the reliance on labeled samples. These approaches typically label samples from a single modality to alleviate the challenges of cross-modal labeling. Unsupervised approaches \cite{8,9,10,44} operate under the assumption that both intra-modal and inter-modal samples are unlabeled. Effective training requires accurate and reliable pseudo-label prediction to enable supervised learning. However, existing pseudo-label prediction methods are sensitive to the quantity of cross-modal positive samples and modality discrepancies, often leading to predictions with many noisy labels, which ultimately hinder model performance enhancement.

In response to the aforementioned issues, this paper proposes a novel weakly supervised paradigm for VIReID. This paradigm relies solely on intra-modal identity labels, eliminating the need for cross-modal identity correspondence annotations, thereby substantially mitigating the challenges and costs associated with cross-modal labeling. As illustrated in Figure \ref{fig1}, this design is driven by two key challenges: First, infrared and visible cameras often operate at different times, making the direct establishment of cross-modal pedestrian correspondences challenging in real-world scenarios; second, the two modalities differ significantly in spectral characteristics and visual appearance, rendering manual cross-modal identity annotation both time-consuming and labor-intensive. In contrast, acquiring identity annotations within a single modality is comparatively easier. Therefore, by fully exploiting intra-modal identity labels, this paper indirectly infers cross-modal identity correspondences. This strategy not only significantly reduces the cost of labeling but also effectively reduces the accumulation of label noise observed in existing methods.

\begin{figure*}[t!]
	\centering
	\includegraphics[height=3.1in]{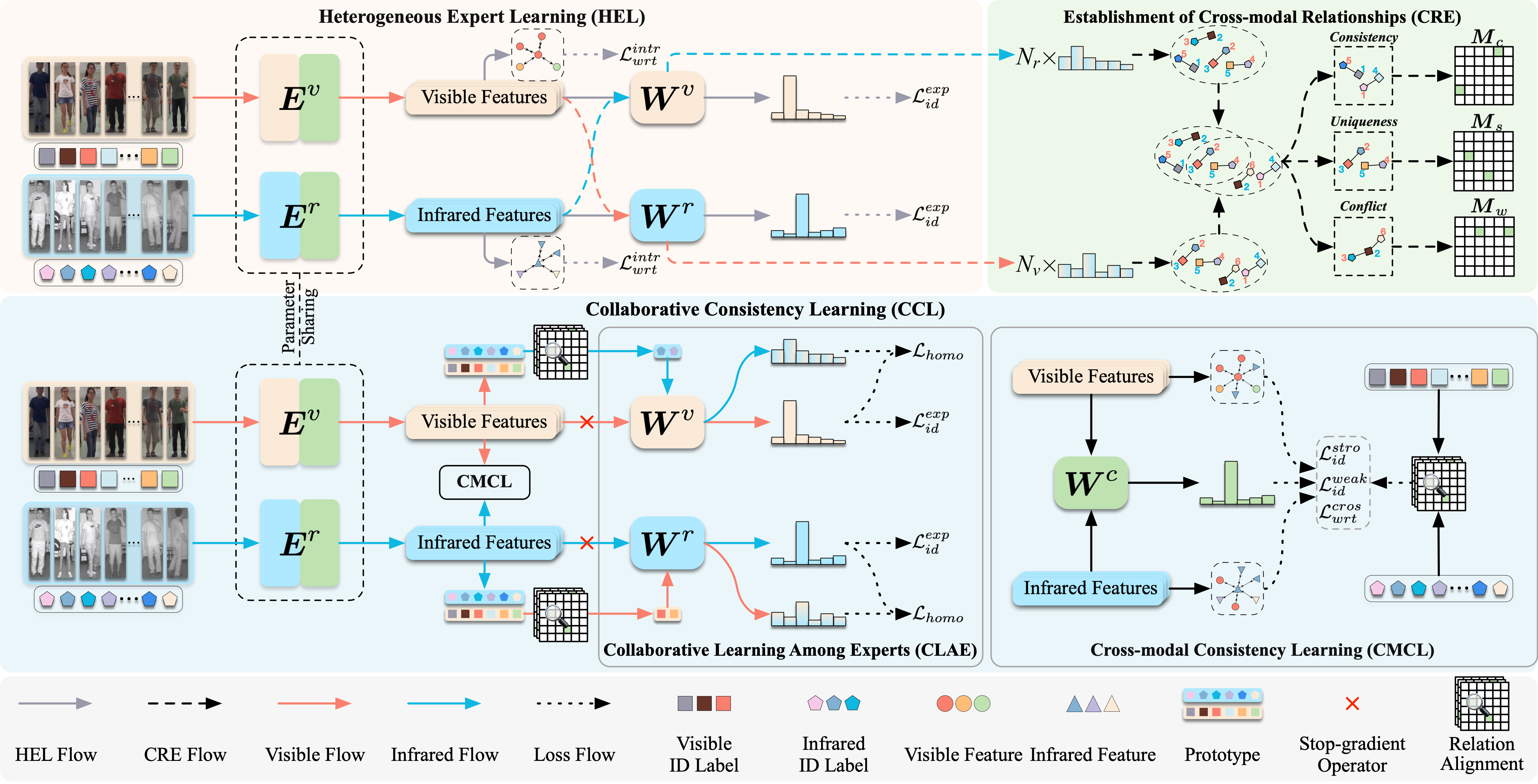}\vspace{-1.5mm}
	\caption{Overview of the proposed method. HEL first performs intra-modality training to obtain experts for each modality, generating cross-modality prediction results for CRE. CRE then establishes robust cross-modality correspondences by fusing predictions from multiple experts. CCL utilizes the correspondences provided by CRE to facilitate CMCL and CLAE.}\vspace{-4.3mm}
	\vspace{-2mm}
	\label{fig2}
\end{figure*}

To establish robust cross-modal identity correspondences in a weakly supervised environment, this paper proposes a heterogeneous expert collaborative consistency learning framework. This framework leverages labeled samples within each modality to independently train dedicated classification experts. These classification experts serve as heterogeneous predictors, estimating pedestrian identities for cross-modal samples. Specifically, they predict and output the identities of cross-modal samples. To enhance prediction accuracy, we propose a cross-modal relationship fusion mechanism that integrates the predictions from both experts, forming accurate inter-modal identity correspondences. Guided by these correspondences, the model gradually enhances its modality-invariant feature representation through iterative optimization and collaborative consistency learning among experts. Throughout this process, the experts and the model mutually reinforce each other, effectively tackling the challenges of training without cross-modal identity correspondences.  The main innovative contributions of this paper are as follows:
\begin{itemize}
	\item We explore VIReID under a weakly supervised setting for the first time and propose an effective mechanism for predicting cross-modal sample relationships. Our method requires annotations only for single-modal samples, eliminating the need for explicit cross-modal correspondence annotations. This avoids labor-intensive process of cross-modal annotation, significantly reducing annotation costs.
	
	\item We design a heterogeneous expert collaborative consistency learning framework. This framework constructs reliable cross-modal identity correspondences by integrating predictions from experts, enabling supervised refinement of the ReID model. Furthermore, by dynamically guiding experts to make consistent identity predictions within and across modalities, our framework achieves collaborative and mutually reinforcing training between experts and the model.
	
	\item We conduct extensive experimental validation on the SYSU-MM01, and LLCM datasets. Experimental results consistently validate the effectiveness of our method, achieving performance comparable to certain fully supervised approaches.
\end{itemize}
\vspace{-1.5mm}

\section{Related Work}
\subsection{Supervised VIReID} 
The primary challenge in supervised VIReID is to overcome cross-modal feature mismatch caused by modality discrepancies. To address this issue, researchers have proposed various effective methods, which can be categorized into three types based on their strategies for handling modality differences \cite{57}: modality-shared representation learning (MSRL) methods \cite{1,12,13,14}, modality-specific information compensation (MSIC) methods \cite{15,16}, and external information-assisted (EIA) methods \cite{17,18,19}. MSRL methods aim to extract common features from both VIS-IR pedestrian images to mitigate the modality gap. Specifically, these methods can be classified into three categories: feature mapping \cite{20,21,3,22}, which aligns the representations of IR-VIS images in a shared feature space; feature disentanglement \cite{23,24,25,26,27}, which often leverages adversarial learning to separate modality-invariant features while preserving identity-related information; and metric learning \cite{4,28,29,30,31}, which imposes identity consistency constraints to directly learn representations that are both identity-discriminative and modality-invariant. However, some of these methods may inadvertently lead to a loss of modality-specific identity-related information, ultimately weakening identity discrimination.

To mitigate this issue, MSIC methods \cite{15,16,33} generate missing modality information so that cross-modal samples of the same identity share consistent representations. Meanwhile, EIA methods leverage external cues (e.g., pedestrian poses \cite{34,35} and masks \cite{36,37,38}) to enhance cross-modal consistency and reduce the impact of modality differences on identification. In a supervised setting, the success of these methods often depends on large-scale labeled training data. However, due to modality differences, labeling extensive cross-modal datasets requires significant human resources. Thus, developing more cost-effective training paradigms is crucial.

\subsection{Semi-supervised VIReID}
To reduce dependence on cross-modal labeled samples, researchers have proposed semi-supervised VIReID methods. These methods typically leverage labeled visible samples and unlabeled infrared samples for training. However, a key challenge lies in effectively establishing correspondences between samples from different modalities. To address this, the OTLA \cite{5} method introduces an optimal transport strategy, modeling label correspondences as an optimal transport problem. However, OTLA may still introduce label noise in predictions. To mitigate this issue, DPIS \cite{6} and MUCG \cite{7} reweight the loss function based on pseudo-label confidence, reducing the adverse impact of pseudo-label noise on model performance. Nevertheless, the absence of infrared sample labels in semi-supervised methods increases the risk of label noise in cross-modal samples, thereby limiting further performance gains.

\subsection{Unsupervised VIReID}
To eliminate reliance on labeled samples, unsupervised VIReID methods have been proposed. These methods \cite{58,55} generate pseudo-labels via clustering within the same modality and establish cross-modal correspondences using similarity-based matching. However, cross-modal pseudo-label noise limits the performance of these methods. To address this, researchers have explored various strategies. Latent sample mining \cite{42,8,9,10} expands the search space for pseudo-label correspondences, reducing mismatches. Auxiliary information methods \cite{43,44,45}  use intermediate modality information to bridge and enhance cross-modal matching. Multi-level difference optimization \cite{46,47,48,49} decomposes cross-modal differences into intra- and inter-modality components, gradually improving representation consistency and reducing label noise. Cross-modal relationship integration \cite{59,60,56,61,62} refines pseudo-labels by aligning multiple correspondences. Despite these efforts, intra-modality label noise still affects inter-modality matching, which limits performance gains. This study explores VIReID under a weakly supervised setting, addressing cross-modal labeling challenges in real-world scenarios while balancing annotation cost and model effectiveness.

\vspace{-1.5mm}

\section{Methodology}
\label{sec:formatting}
\subsection{Problem Definition and Method Overview}
Given a dataset $ \mathcal{X} = \{ \mathcal{X}^v, \mathcal{X}^r \} $ comprising visible and infrared person images, where $ \mathcal{X}^v = \{ x_i^v \}_{i = 1}^{N^v} $ and $ \mathcal{X}^r = \{ x_i^r \}_{i = 1}^{N^r} $ denote the sets of $ N^v $ visible and $ N^r $ infrared person images, respectively. The corresponding label sets are given by $ \mathcal{Y}^v = \{ \bm{y}_i^v \}_{i = 1}^{N^v} $ and $ \mathcal{Y}^r = \{ \bm{y}_i^r \}_{i = 1}^{N^r} $, where $ \bm{y}_i^v \in \mathbb{R}^{1\times C^v} $ and $ \bm{y}_i^r \in \mathbb{R}^{1\times C^r} $ represent the labels associated with samples $ x_i^v $ and $ x_i^r $, respectively. Under the weakly supervised setting considered in this paper, label information is available within each modality, whereas cross-modal label correspondences remain unknown.

To address this issue, we propose the method illustrated in Figure~\ref{fig2}, which consists of heterogeneous expert learning (HEL), cross-modal relationship establishment (CRE), and collaborative consistency learning (CCL). During training, single-modality classifiers serve as heterogeneous experts to learn cross-modal consistency in feature representation. Cross-modal relationships are established by comparing the consistency of predictions made by these experts. Based on the correspondences, prototypes of the same identity across modalities are selected for collaborative learning among the experts. Meanwhile, leveraging these inter-modal correspondences, the encoder and shared classifier are further updated, enabling the network to extract features that maintain cross-modal consistency for the same identity.

\subsection{Heterogeneous Expert Learning}
To establish identity correspondences between cross-modal images, we propose a cross-modality judgment approach. Specifically, to fully exploit identity-discriminative features within each modality, we construct separate pedestrian identity classifiers, denoted as $ \bm{W}^v $ and $ \bm{W}^r $, for visible and infrared images, respectively. Each classifier is trained under supervision from identities within its respective modality, enabling the feature encoders to learn identity-discriminative representations. As shown in Figure~\ref{fig2}, both feature encoders, $ \bm{E}^v $ and $ \bm{E}^r $, adopt ResNet-50 \cite{51} as the backbone. The first layers of each encoder serve as a modality-specific encoder with unshared parameters, while the remaining convolutional layers are shared to enhance cross-modal feature consistency.

Given the training samples $x_i^v$ and $x_i^r$, we pass them through the encoders $\bm{E}^v$ and $\bm{E}^r$ to obtain the feature representations $\bm{f}_i^v$ and $\bm{f}_i^r$. To enhance their discriminative capability, we employ a combination of cross-entropy loss and weighted regularization triplet loss \cite{13} to update $\bm{E}^v$ and $\bm{E}^r$. The cross-entropy loss is defined as:
\begin{equation}\small
	\mathcal{L}_{id}^{exp} = -\sum_{t \in \{v,r\}}\frac{1}{n^t} \sum_{i=1}^{n^t} \bm{y}_i^t \log \bm{p}_i^t,
\end{equation}
where $n^t$ denotes the number of samples of modality $t$ in a mini-batch, and $\bm{p}_i^t = \bm{W}^t(\bm{f}_i^t) \in \mathbb{R}^{C_t \times 1}$ represents the identity prediction vector produced by the classifier $\bm{W}^t$ (for $t = v, r$) given the features $\bm{f}_i^v$ and $\bm{f}_i^r$.  

The weighted regularization triplet loss is defined as:
\begin{equation}\small
	\mathcal{L}_{wrt}^{intr} = \sum_{t \in \{v,r\}} \frac{1}{n^t} \sum_{i=1}^{n^t} \log( 1 + \exp( \sum_j w_{ij}^p d_{ij}^p - \sum_k w_{ik}^n d_{ik}^n)) 
\end{equation}
where
\begin{equation}\small
	w_{ij}^p = \frac{\exp(d_{ij}^p)}{\sum_{{d_{ij}} \in {\cal P}_i} \exp(d_{ij}^p)}, \quad w_{ik}^n = \frac{\exp(-d_{ik}^n)}{\sum_{{d_{ik}} \in {\cal N}_i} \exp(-d_{ik}^n)},
\end{equation}
with ${\cal P}_i$ and ${\cal N}_i$ denoting the sets of positive and negative samples, respectively, for the anchor sample $x_i^t$ within the same modality in a mini-batch. Here, $j$ and $k$ index the positive and negative samples, while $d_{ij}^p = \|\bm{f}_i^t - \bm{f}_j^t\|_2$ and $d_{ik}^n = \|\bm{f}_i^t - \bm{f}_k^t\|_2$ represent the Euclidean distances between positive and negative pairs, respectively. The classifiers $\bm{W}^v$ and $\bm{W}^r$, trained by minimizing the loss functions in Eqs. (1) and (2), serve as experts in identity discrimination for their respective modalities. Since $\bm{W}^v$ and $\bm{W}^r$ are trained on distinct modalities and focus on different identity-related information in pedestrian samples, we refer to them as heterogeneous experts.

\subsection{Establishment of Cross-Modal Relationships}
Although the encoders trained with the constraints in Eq.(1) and Eq.(2) extract features with certain discriminative power, the absence of cross-modality sample pairs poses challenges for $\bm{W}^v$ and $\bm{W}^r$ in cross-identification. To address this issue, it is essential to establish correspondences between modalities. To this end, we propose a cross-modality identity relationship establishment mechanism. This mechanism mitigates the limitations of individual experts by integrating their predictions to construct reliable cross-modality identity correspondences. Specifically, after training the heterogeneous experts using the loss functions in Eq.(1) and Eq.(2), these experts are employed to predict the identities of all samples from the other modality. In the classification vector, the score at each position represents the probability that a given sample belongs to the corresponding identity in the other modality.

Similar to \cite{55}, we employ the Count Priority Selection method to obtain the expert decision matrix $\bm{M}^{t \to \bar{t}}$, where $t, \bar{t} \in \{v, r\}, t \neq \bar{t}$. The notation $t \to \bar{t}$ denotes the prediction results of $\bar{t}$-modality experts on $t$-modality samples, and $\bm{M}^{t \to \bar{t}} \in \mathbb{R}^{C^{t} \times C^{\bar{t}}}$ encodes the identity correspondences between the two modalities. For example, the element at the $i$-th row and $j$-th column of $\bm{M}^{v \to r}$ represents the correspondence between the $i$-th visible pedestrian identity and the $j$-th infrared pedestrian identity. If the two identities correspond, the respective element in $\bm{M}^{v \to r}$ is set to 1; otherwise, it is 0. While relying solely on identical predictions across experts as pseudo-labels ensures high reliability, it may overlook many potential identity correspondences. To mitigate this issue, we classify all cross-modal identity correspondences into three categories based on the consistency of expert predictions: consistent matches, unique matches, and conflicting matches.

Based on the definition of consistent matching, the consistent identity relationship matrix $\bm{M}_c$ is obtained as follows:
\begin{equation} \small \bm{M}_c = \bm{M}^{t \to \bar{t}} \odot( \bm{M}^{\bar{t} \to t})^{\rm{T}} \end{equation}
where $\odot$ denotes the Hadamard product. When $\bm{M}_c(i,j) = 1$, it indicates that the experts unanimously agree that the $i$-th identity in the $t$ modality corresponds to the $j$-th identity in the $\bar{t}$ modality.
The single identity relationship matrix for one expert is defined as:
\begin{equation} \small
	\begin{split}
		\bm{M}_s^{t \to \bar{t}} &= \bm{A}^{\rm{T}} \odot \bm{M}^{t \to \bar{t}}, \quad \bm{A}(i,j) = r_i \cdot c_j, \\
		r_i &= \bm{1}_{\{0\}} ( \sum_{j=1}^{C^{t}} \bm{M}^{\bar{t} \to t}(i,j)), \\
		c_j &= \bm{1}_{\{0\}} ( \sum_{i=1}^{C^{\bar t}} \bm{M}^{\bar{t} \to t}(i,j)),
	\end{split}
\end{equation}
where $\bm{1}_{\{0\}}(\cdot)$ is an indicator function that returns 1 if the input equals 0 and 0 otherwise.
Let $\bm{M}_s = \bm{M}_s^{v \to r} + \left( \bm{M}_s^{r \to v} \right)^{\rm{T}}$. The contradictory identity relationship matrix is then given by:
\begin{equation} 
	\small \bm{M}_w = \bm{M}^{v \to r} + \left( \bm{M}^{r \to v} \right)^{\rm{T}} - 2\bm{M}_c - \bm{M}_s \end{equation}
The matrices $\bm{M}_c$, $\bm{M}_s$, and $\bm{M}_w$ effectively represent the cross-modal identity correspondences. Consequently, they are used to guide $\bm{E}^v$ and $\bm{E}^r$ to extract consistent features for cross-modal samples of the same identity.

\subsection{Collaborative Consistency Learning}
\textbf{Cross-modal Consistency Learning}. Due to the challenge of unknown pedestrian identities across modal samples, it is not feasible to utilize identity relationships between cross-modal samples to guide the training of encoders, limiting the model's cross-modal matching performance. To address this issue, we propose a cross-modal consistency learning method based on the identity relationships established by $\bm{M}_c$, $\bm{M}_s$, and $\bm{M}_w$ among samples across modalities. In this method, for samples with established identity relationships in $\bm{M}_c$, $\bm{M}_s$, and $\bm{M}_w$, we apply a cross-modal cross-entropy loss as a constraint: 
\begin{equation}\small
	{\cal L}_{id}^{stro} =  - \frac{1}{{n^s}} \sum_{i=1}^{n^s} \hat{\bm{q}}_i^v \log \left( \bm{W}^c(\bm{f}_i^v) \right) - \frac{1}{{n^r}} \sum_{i=1}^{n^r} \bm{q}_i^r \log \left( \bm{W}^c(\bm{f}_i^r) \right)
\end{equation} 
where $n^r$ denotes the number of infrared samples in a mini-batch, and $n^s$ denotes the number of visible samples in the mini-batch that have corresponding identities with the infrared modality samples in the training set. $\hat{\bm{q}}_i^v$ represents the one-hot vector of the infrared modality identity corresponding to the $i$-th visible sample in $\bm{M}_c$ or $\bm{M}_s$—i.e., the pseudo-label for the visible sample. $\bm{W}^c(\bm{f}_i^t)$ represents the classification result of $\bm{f}_i^t$ by the shared classifier $\bm{W}^c$.

For samples in $\bm{M}_w$ with contradictory identity correspondences, we employ a relaxed identity discrimination constraint during model training to mitigate the impact of cross-modal label noise: 
\begin{equation}\small
	{{\cal L}_{id}^{weak}} =  - \frac{1}{n_w^v} \sum_{i = 1}^{n_w^v} \bm{m}_i \log \left( 1 - {\bm{W}^c}(\bm{f}_i^v) + \epsilon \right)
\end{equation}		
\begin{equation}\small		
	m_{il} = \left\{ 
	\begin{array}{l}
		1, \quad \text{if } p_{il}^{v} < \min_{k \in K} \left( p_{ik}^{v} \right), \\
		0, \quad \text{otherwise}
	\end{array}
	\right.
\end{equation}
where $K = \{ k \mid \bm{M}_{w}(j,k) = 1 \}$, $j$ is the intra-modal identity corresponding to sample $x_i^v$, and $\epsilon = 10^{-10}$ is used to prevent numerical overflow. $p_{il}^v$ represents the probability score at the $l$-th position in the classification result ${\bm{W}^c}(\bm{f}_{i}^v)$ for sample $x_i^v$. $m_{il}$ is the $l$-th element of the vector ${\bm{m}_i}$. This loss function effectively leverages samples with uncertain cross-modal identities for model training. Additionally, during model training, we adopt a cross-modal weighted regularization triplet loss, ${\cal L}_{wrt}^{cros}$. Unlike ${\cal P}_i$ in ${\cal L}_{wrt}^{intr}$, ${\cal P}_i$ in ${\cal L}_{wrt}^{cros}$ represents the set of positive samples corresponding to the $i$-th anchor sample $x_i^t$ in a mini-batch, which includes both intra-modal positive samples and cross-modal positive samples obtained from ${\bm{M}}_c$.		
	
\noindent \textbf{Collaborative Learning Among Experts}.
Since experts within each modality are trained on uni-modal samples, their ability to discriminate cross-modal identities is limited. Consequently, when heterogeneous experts predict pedestrian identities across modalities, they may produce inconsistent results for challenging samples, reducing the accuracy of $\bm{M}_c$, $\bm{M}_s$, and $\bm{M}_w$. To address this issue, we propose a collaborative consistency learning method among heterogeneous experts. Specifically, we guide the experts to generate consistent predictions for cross-modal positive samples, introducing a degree of homogeneity to enhance cross-modal consistency. However, within a mini-batch, cross-modal positive samples of the same identity may be absent. Even when present, they may not fully capture the diversity of pedestrian appearances. To mitigate this, we construct prototypes $\mathcal{P}^t = [\mathcal{P}_1^t, \mathcal{P}_2^t, \dots, \mathcal{P}_{C^t}^t]^\top \in \mathbb{R}^{C^t \times d}$ for pedestrian identities in each modality, where $d$ denotes the feature dimension. During training, we initialize the prototypes using the mean feature of each identity and update them as follows:
\begin{equation}\small
	\mathcal{P}_i^t \leftarrow \lambda \mathcal{P}_i^t + (1 - \lambda) \bar{\bm{f}}_i^t
\end{equation}
where $\mathcal{P}_i^t$ represents the prototype feature of the $i$-th identity in modality $t$, $\lambda \in [0,1]$ is the momentum update coefficient, and $\bar{\bm{f}}_i^t$ denotes the mean feature of the samples associated with identity $i$ within a mini-batch.

Assume that $\bm{p}_i^{v \to v} = {\bm{W}^v}(\bm{f}_i^v)$ represents the classification result of $\bm{W}^v$ for $\bm{f}_i^v$, and $\bm{p}_i^{r \to v} = {\bm{W}^v}({\cal \bm{P}}_{\hat y_{i}^r}^r)$ represents the classification result of $\bm{W}^v$ for ${\cal P}_{\hat y_i^r}^r$, where ${\hat y_i^r}$ represents the identity of $x_{i}^v$ in $\bm{M}_c$. To promote consistency in the decision-making process of experts across different modalities, we propose a collaborative consistency loss for $\bm{W}^v$:
\begin{equation}\small
	{\cal L}_{homo}^v = \frac{1}{{n^c} \times C^v} \sum_{i=1}^{n^c} \|\bm{p}_i^{v \to v} - \bm{p}_i^{r \to v}\|_2^2 
\end{equation}
where $n^c$ denotes the number of visible samples in a mini-batch that have identity relationships in ${\bm{M}}_c$. Similarly, we define the collaborative consistency loss ${\cal L}_{homo}^r$ for ${\bm{W}^r}$. It is important to note that the cross-modal identity relationships predicted by experts may not always be accurate, and directly enforcing a closer match between the two modalities could negatively impact model training.

Typically, when an expert's prediction confidence is high, the probability assigned to the predicted identity is large, whereas when the prediction is uncertain, this probability tends to be smaller. Based on this observation, we use a weight derived from the information entropy $H$ of the prediction results to dynamically adjust the constraint strength in the expert homogeneity loss. The more reliable an expert's prediction, the weaker the constraining effect of the loss. Specifically, the total collaborative consistency loss is formulated as follows:
\begin{equation}\small
	{\cal L}_{homo} = w_v {\cal L}_{homo}^v + w_r {\cal L}_{homo}^r
\end{equation}
where
\begin{equation}\resizebox{0.88\hsize}{!}{$
		w_v = \frac{H(\bm{p}^{r \to v})}{H(\bm{p}^{r \to v}) + H(\bm{p}^{v \to r})}, w_r = \frac{H(\bm{p}^{v \to r})}{H(\bm{p}^{r \to v}) + H(\bm{p}^{v \to r})}$}
\end{equation}
Additionally, to prevent experts from forgetting their original knowledge, we continue training them within their respective modalities by minimizing the loss in Eq. (1).

\subsection{Model Optimization}
In the expert construction phase, the overall optimization objective is formulated as:
\begin{equation}\small
	{{\cal L}_{phase 1}} = {\cal L}_{id}^{exp} + {\lambda_1}{\cal L}_{wrt}^{intr} 
\end{equation}
where ${\lambda_1}$ is a hyperparameter used to balance the model.

In collaborative consistency learning phase, the overall optimization objective is:
\begin{equation}\resizebox{0.88\hsize}{!}{$
		{{\cal L}_{phase 2}} = {\cal L}_{id}^{exp} + {\cal L}_{id}^{stro} + {\cal L}_{homo}  + {\lambda_1}{\cal L}_{wrt}^{cros}+ {\lambda_2}{\cal L}_{id}^{weak}$}
\end{equation}
where ${\cal L}_{wrt}^{cros}$ is the weighted regularization triplet loss between cross-modal samples, as defined in Eq. (2). During collaborative consistency learning, cross-modal consistency learning and expert collaborative learning reinforce each other, enabling the network to extract cross-modal consistent features and effectively addressing the absence of cross-modal sample labels. Simultaneously, heterogeneous experts provide more reliable identity predictions during the homogenization process, significantly improving the accuracy of $\bm{M}_c$, $\bm{M}_s$, and $\bm{M}_w$.

\begin{table*}[t!]
	\begin{center}
		\caption{\small Comparison of CMC (\%), mAP (\%) and mINP (\%) performances with the state-of-the-art methods on \textbf{SYSU-MM01} dataset.}
		\vspace{-8pt}
		\resizebox{0.9\linewidth}{!}{
			\begin{tabular}{c|c|c||ccc|cc||ccc|cc}    
				\bottomrule	 
				\multicolumn{3}{c||}{SYSU-MM01 Settings}&\multicolumn{5}{c||}{All Search}&\multicolumn{5}{c}{Indoor Search}\\
				\hline
				Type & Methods&Venue& Rank-1 & Rank-10 & Rank-20 & mAP & mINP&Rank-1 & Rank-10 & Rank-20 & mAP &mINP\\
				\bottomrule	
								\multirow{6}{*}{SVIReID}
				& AGW\cite{13} & TPAMI'21 & 47.5 & 84.4 & 92.1 & 47.7 & 35.3 & 54.2 & 91.1 & 95.6 & 63.0 & 59.2 \\
				& LbA\cite{1} & ICCV'21 & 55.4 & - & - & 54.1 & - & 58.5 & - & - & 66.3 & - \\
				& CAJ\cite{14} & ICCV'21 & 69.9 & 95.7 & 98.5 & 66.9 & 53.6 & 76.3 & 97.9 & 99.5 & 80.4 & 76.8 \\
				& DEEN\cite{3} & CVPR'23 & 74.7 & 97.6 & 99.2 & 71.8 & - & 80.3 & 99.0 & 99.8 & 83.3 & - \\
				& MUN \cite{52} & ICCV'23 & 76.2 & 97.8 & - & 73.8 & - & 79.4 & 98.1 & - & 82.1 & - \\
				& CSDN \cite{53} & TMM'25 & 76.7 & 97.2 & 99.1 & 73.0 & - & 84.5 & 99.2 & 99.6 & 86.8 & - \\
				\midrule \multirow{4}{*}{SSVIReID}
				& MAUM-50 \cite{54} & CVPR'22 & 28.8 & - & - & 36.1 & - & - & - & - & - & - \\
				& MAUM-100 \cite{54} & CVPR'22 & 38.5 & - & - & 39.2 & - & - & - & - & - & - \\
				& OTLA \cite{5} & ECCV'22 & 48.2 & - & - & 43.9 & - & 47.4 & - & - & 56.8 & - \\
				& DPIS \cite{6} & ICCV'23 & 58.4 & - & - & 55.6 & - & 63.0 & - & - & 70.0 & - \\
				\midrule \multirow{6}{*}{USVIReID}
				& ADCA \cite{55} & MM'22 & 45.5 & 85.3 & 93.2 & 42.7 & 28.3 & 50.6 & 89.7 & 96.2 & 59.1 & 55.2 \\
				& CCLNet \cite{43} & MM'23 & 54.0 & 88.8 & 95.0 & 50.2 & - & 56.7 & 91.1 & 97.2 & 65.1 & - \\
				& GUR \cite{46} & ICCV'23 & \underline{63.5} & - & - & \underline{61.6} & \underline{47.9} & \underline{71.1} & - & - & \underline{76.2} & \underline{72.6} \\
				& MMM \cite{48} & ECCV'24 & 61.6 & 93.3 & \underline{98.0} & 57.9 & - & 64.4 & 95.0 & 98.2 & 70.4 & - \\
				& DLM \cite{56} & TPAMI'25 & 62.2 & \underline{93.4} & 97.4 & 58.4 & 43.7 & 67.3 & \underline{95.9} & \underline{99.2} & 72.7 & 68.9 \\
				& PCAL \cite{8} & TIFS'25 & 57.9 & 91.0 & 96.4 & 52.9 & 36.9 & 60.1 & 93.3 & 97.3 & 66.7 & 62.1 \\
				\midrule
				WSVIReID & Ours & - & \textbf{70.4} & \textbf{95.8} & \textbf{98.8} & \textbf{66.6} & \textbf{52.7} & \textbf{76.5} & \textbf{97.5} & \textbf{99.3} & \textbf{80.2} & \textbf{76.6} \\
				\bottomrule
		\end{tabular}		}
		\vspace{-6.5mm}
		\label{tab1}
	\end{center}
\end{table*}

\section{Experiments}
\subsection{Datasets and Evaluation Protocols}
To evaluate the proposed method, we assess its performance on two public VIReID datasets: SYSU-MM01 \cite{20} and LLCM \cite{3}.
\begin{table}[t!]
	\begin{center}
		\caption{\small Comparison of Rank-1 (\%) and mAP (\%) performances with the state-of-the-art methods on \textbf{LLCM} dataset, where * denotes the results obtained using publicly available code for the experiment.}
		\vspace{-8pt}
		\resizebox{\linewidth}{!}{
			\begin{tabular}{c|c|c|cc|cc}
				\bottomrule
				\multicolumn{3}{c|}{LLCM Settings}&\multicolumn{2}{c|}{VIS to IR}&\multicolumn{2}{c}{IR to VIS}\\
				\hline
				Type&Methods&Venue&Rank-1&mAP&Rank-1&mAP\\
				\bottomrule
				\multirow{5}{*}{SVIReID}
				& AGW \cite{13} & TPAMI'21 & 51.5 & 55.3 & 43.6 & 51.8 \\
				& LbA\cite{1} & ICCV'21 & 50.8 & 55.6 & 43.8 & 53.1 \\
				& CAJ \cite{14} & ICCV'21 & 56.5 & 59.8 & 48.8 & 56.6 \\
				& DEEN \cite{3} & CVPR'23 & 62.5 & 65.8 & 54.9 & 62.9 \\
				& CSDN \cite{53} & TMM'25 & 63.7 & 66.5 & 55.8 & 63.5 \\
				\midrule
				\multirow{1}{*}{SSVIReID}
				& OTLA \cite{5} & ECCV'22 & 44.2 & 48.2 & 36.2 & 42.2 \\
				\midrule
				\multirow{2}{*}{USVIReID} &ADCA$^*$\cite{55} &MM'22&42.5&46.9&38.4&44.4\\
				&PGM$^*$\cite{42}&CVPR'23&\underline{44.9}&\underline{49.0}&\underline{39.4}&\underline{45.3}\\
				\midrule
				WSVIReID & Ours & - & \textbf{55.3} & \textbf{58.7}& \textbf{47.3} & \textbf{53.3} \\
				\bottomrule
			\end{tabular}
		}
	   \vspace{-7.5mm}
		\label{tab2}
	\end{center}
\end{table}

\noindent\textbf{SYSU-MM01}, the first large-scale VIReID dataset, contains images captured by four visible cameras and two near-infrared cameras across various indoor and outdoor scenes. It includes 491 pedestrian identities, with a total of 287,628 visible images and 15,792 infrared images. The training set consists of 22,258 visible images and 11,909 infrared images from 395 identities, while the test set covers 96 identities. We evaluate performance under both all-search and indoor-search modes.

\noindent\textbf{LLCM}, one of the most challenging large-scale VIReID datasets, comprises images captured by nine visible cameras and nine infrared cameras deployed in low-light environments. It contains 46,767 images from 1,064 identities, including 16,946 visible and 13,975 infrared images. The training set includes samples from 713 identities, while the test set consists of 8,680 visible and 7,166 infrared images from 351 identities. We assess performance in both visible-to-infrared and infrared-to-visible retrieval modes.

\noindent\textbf{Evaluation Protocols}. Following standard practice, we use Cumulative Matching Characteristic (CMC) \cite{29}, mean Average Precision (mAP), and mean Inverse Negative Penalty (mINP) \cite{13} for evaluation.

\subsection{Implementation Details}
The proposed method is implemented in PyTorch, with experiments conducted on a single RTX 4090 GPU. ResNet-50 serves as the encoder, and all input images are resized to $288 \times 144$. During training, data augmentation techniques, including style augmentation \cite{63}, random cropping, horizontal flipping, and random erasing, are applied. After completing heterogeneous expert learning, collaborative consistency learning is performed for 120 epochs. The initial learning rate is set to $3 \times 10^{-4}$ for the encoder and $6 \times 10^{-4}$ for the experts and shared classifier. A warmup strategy is applied for the first 10 epochs to gradually increase the learning rate, which is then decayed to 10\% at the 30th and 70th epochs. The hyper-parameter $\lambda$ for prototypes is set to 0.8, while $\lambda_1$ and $\lambda_2$ are both set to 0.25.

\subsection{Comparasion with State-of-the-Art Methods}
To validate the effectiveness of our method, we compare it with state-of-the-art approaches in supervised VIReID (SVIReID), semi-supervised VIReID (SSVIReID), and unsupervised VIReID (USVIReID). The experimental results on the SYSU-MM01 and LLCM datasets are presented in Tables \ref{tab1} and \ref{tab2}, with the best performance highlighted in bold and the second-best underlined.

\noindent\textbf{Comparison with Supervised Methods}. In a weakly supervised setting with missing cross-modal annotations, our method achieves performance on the SYSU-MM01 and LLCM datasets that is comparable to or even surpasses some SVIReID methods. This demonstrates its ability to effectively handle missing cross-modal annotations and learn cross-modal invariant features.

\noindent\textbf{Comparison with Semi-Supervised Methods}. As shown in Tables \ref{tab1} and \ref{tab2}, our method significantly outperforms existing SSVIReID approaches. As previously noted, pseudo-label noise caused by missing infrared annotations hinders the performance of semi-supervised methods, especially on the challenging SYSU-MM01 and LLCM datasets. Specifically, on SYSU-MM01 (all search), our method improves Rank-1 accuracy and mAP by 12.0\% and 11.0\%, respectively, compared to DPIS. It also significantly outperforms MAUM-50 and MAUM-100, which are trained with 50 and 100 infrared identities, respectively. On LLCM (VIS to IR), our method achieves improvements of 11.1\% and 10.5\% in Rank-1 accuracy and mAP, respectively, over OTLA. These results validate its effectiveness in addressing the challenge of missing cross-modal labels.

\noindent\textbf{Comparison with Unsupervised Methods}. Compared to the USVIReID methods, our approach achieves substantial performance gains with minimal annotation costs. Specifically, on SYSU-MM01 (all search), it surpasses GUR by 6.9\% and 5.0\% in Rank-1 accuracy and mAP, respectively. On LLCM (VIS to IR), it outperforms PGM by 10.4\% and 9.7\% in Rank-1 accuracy and mAP, respectively. These improvements stem from the collaborative promotion mechanism, which effectively establishes relationships between cross-modal samples and enhances supervision.

\subsection{Ablation Study}
The proposed method consists of three main components: HEL, CRE, and CCL. The model trained solely with HEL serves as the Baseline. HEL enables the model to acquire fundamental cross-modal recognition capabilities, providing a foundation for constructing cross-modal correspondences based on prediction results. CRE is designed to enhance cross-modal correspondences for CMCL and must be evaluated in conjunction with CMCL. To assess its impact, we compare consistency prediction using heterogeneous experts as cross-modal correspondences with CRE+CMCL results.
CCL comprises two parts: CMCL and CLAE. In the ablation study, CMCL without CRE prediction results is denoted as B+CMCL\textbackslash CRE, while incorporating CRE is denoted as B+CRE+CMCL. The full model is referred to as B+CRE+CMCL+CLAE. Table \ref{tab3} presents the experimental results for different components.

\noindent\textbf{Effectiveness of HEL}. HEL is a core component of the Baseline model. On the SYSU-MM01 dataset, the Baseline achieves Rank-1 and mAP scores of 47.8\% and 47.2\%, respectively, in all-search mode, and 60.8\% and 67.7\%, respectively, in indoor-search mode. These results indicate that a model trained solely within modalities already possesses some cross-modal retrieval capability, providing a strong foundation for further optimization. However, intra-modal training is limited to optimizing within each modality and does not establish correspondences between cross-modal samples. Therefore, relying solely on intra-modal training is insufficient to address the challenges posed by the absence of inter-modal correspondences in VI-ReID.

\noindent\textbf{Effectiveness of CMCL}. Using only expert-provided consistency predictions for cross-modal consistency learning, B+CMCL\textbackslash CRE improves Rank-1 and mAP by 18.9\% and 15.6\%, respectively, over the Baseline in the all-search setting. In the indoor-search setting, Rank-1 and mAP improve by 12.6\% and 9.9\%, respectively. These results demonstrate that CMCL effectively reduces modality differences and enhances cross-modal retrieval. By leveraging CMCL, the model learns more consistent feature representations across modalities, leading to improved retrieval performance.
\begin{table}[t!]
	\begin{center}
		\caption{\small Ablation study on \textbf{SYSU-MM01}. ``B'' denotes the Baseline method. }
		\vspace{-8pt}
		\resizebox{0.9\linewidth}{!}{
			\begin{tabular}{l|cc|cc}
				\bottomrule
				\multirow{2}{*}{Settings}&\multicolumn{2}{|c}{All Search}&\multicolumn{2}{|c}{Indoor Search}\\
				\cline{2-5}
				&Rank-1&mAP&Rank-1&mAP\\
				\bottomrule
				B&47.8 & 47.2 & 60.8 & 67.7 \\
				B+CMCL\textbackslash CRE&66.7 & 62.8 & 73.4 & 77.6\\
				B+CRE+CMCL&68.0 & 64.6 & 74.0 & 78.3 \\
				B+CRE+CMCL+CLAE & \textbf{70.4} & \textbf{66.6} & \textbf{76.5} & \textbf{80.2}\\
				\bottomrule
		\end{tabular}}\vspace{-7.5mm}
		\label{tab3}
	\end{center}
\end{table}

\noindent\textbf{Effectiveness of CRE}. By introducing the cross-modal relation establishment strategy, B+CRE+CMCL improves Rank-1 and mAP by 1.3\% and 1.8\%, respectively, over B+CMCL\textbackslash CRE in the all-search setting. In the indoor-search setting, Rank-1 and mAP improve by 0.6\% and 0.7\%, respectively. This improvement arises because the relation establishment strategy not only integrates consistent predictions from experts but also leverages inconsistent ones. Consequently, it facilitates more comprehensive cross-modal identity correspondence and effectively mitigates relation-matching errors caused by the limitations of individual experts.

\begin{table}[h]
	\begin{center}
		\caption{\small Further analysis of CRE effectiveness on \textbf{SYSU-MM01}.}
		\vspace{-8pt}
		\resizebox{0.9\linewidth}{!}{
	  \begin{tabular}{l|cc|cc}
		\bottomrule
		\multirow{2}{*}{Settings}&\multicolumn{2}{c|}{All Search}&\multicolumn{2}{c}{Indoor Search}\\
		\cline{2-5}
		&Rank-1&mAP&Rank-1&mAP\\
		\bottomrule
		B+CMCL($\bm{M}^{r\to v}$)&61.0&57.5&69.8&75.1\\
		B+CMCL($\bm{M}^{v\to r}$)&67.7 & 64.4 & 73.7 & 77.8\\
		B+CMCL\textbackslash CRE&66.7 & 62.8 & 73.4 & 77.6\\
		B+CRE+CMCL&\textbf{68.0} & \textbf{64.6} & \textbf{74.0} & \textbf{78.3}\\
		\bottomrule
	  \end{tabular}
	}
	\label{tab5}
	\end{center}
	\vspace{-3.5mm}
\end{table}

\noindent\textbf{Effectiveness of CLAE}. After introducing CLAE, B+CRE+CMCL+CLAE improves Rank-1 and mAP by 2.4\% and 2.0\%, respectively, over B+CRE+CMCL in the all-search setting. In the indoor-search setting, Rank-1 and mAP improve by 2.5\% and 1.9\%, respectively. These results indicate that CLAE further enhances identity prediction accuracy for cross-modal samples. By leveraging CLAE, the two heterogeneous experts progressively reduce modality discrepancies, enhance collaboration, and generate more consistent cross-modal predictions. \textbf{Additional experimental results are in the supplementary material.}

\subsection{Further Analysis of CRE}
To further analyze the role of the proposed CRE, we utilize the cross-modal relationships established by $\bm{M}^{r\to v}$ and $\bm{M}^{v\to r}$ in Eq. (6) to guide model training. The models trained with these relationships are denoted as B+CMCL($\bm{M}^{r\to v}$) and B+CMCL($\bm{M}^{v\to r}$), respectively. As shown in Table~\ref{tab5}, B+CMCL($\bm{M}^{r\to v}$) exhibits significantly lower recognition performance compared to B+CMCL($\bm{M}^{v\to r}$). This is mainly because infrared images lack discriminative color information, limiting the ability of visible expert to recognize infrared images, whereas infrared expert do not face this issue and thus perform better. Since the recognition abilities of different experts are complementary, the B+CMCL model trained solely on consistent predictions among experts fails to fully exploit this complementarity. The CRE method effectively addresses this issue, and with its assistance, the B+CRE+CMCL model achieves the best recognition performance, further validating the effectiveness of CRE.
\vspace{-2.5mm}
\begin{figure}[ht!]
	\centering
	\includegraphics[height=2.3in]{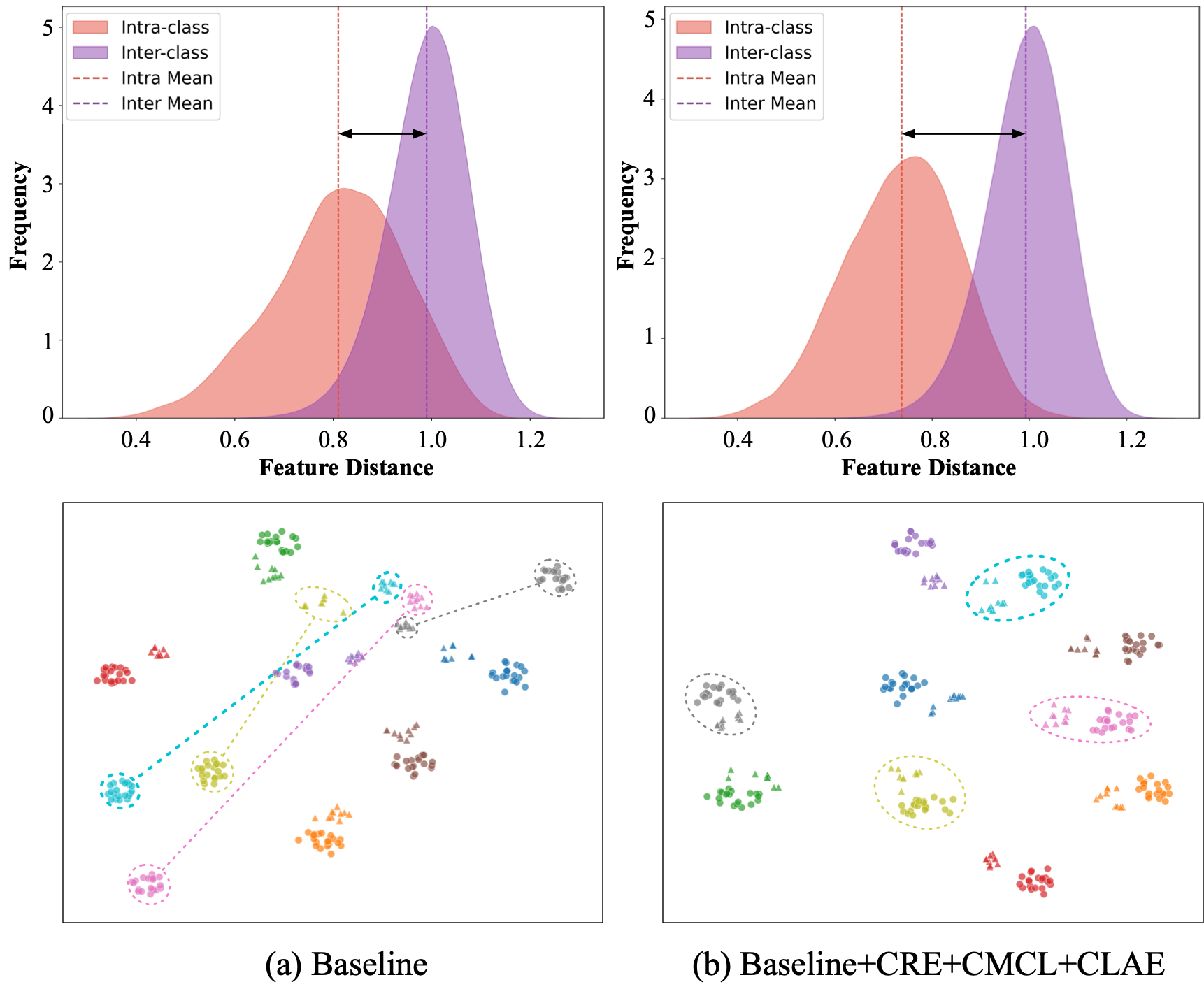}\vspace{-2.5mm}
	\caption{Visualization comparison of feature distributions between baseline and our method on SYSU-MM01: intra- and inter-class distances of cross-modal sample features in the first row, and t-SNE mapped 2D feature space visualization in the second row, with 10 randomly selected identities and 10 samples each.}\vspace{-3.0mm}
	\label{fig6}
\end{figure}

\subsection{Visualization Analysis}

We further validate our method by visualizing feature distributions. Figure~\ref{fig6} (top) compares the intra-class and inter-class distances of features from cross-modal samples between the baseline and our method. Our method significantly increases the gap between intra-class and inter-class distances, effectively reducing the distance between same-identity cross-modal samples and minimizing modality discrepancies. Figure ~\ref{fig6} (bottom) shows the t-SNE visualization of features in 2D, with circular points for visible images, triangular points for infrared images, and colors representing pedestrian identities. Compared to the baseline, our method clusters same-identity features more effectively while maintaining separation between different identities, demonstrating its ability to mitigate modality discrepancies. 
\section{Conclusion}
This paper proposes a weakly supervised cross-modal reID method that establishes identity correspondences without labeled cross-modal identity pairs. By leveraging a heterogeneous expert collaborative consistency learning framework, our method effectively integrates predictions from modality-specific classification experts through a cross-modal relationship fusion mechanism. Experimental results on two challenging datasets validate its effectiveness in enhancing cross-modal identity recognition and extracting modality-invariant representations. This work provides a promising direction for reducing reliance on fully labeled cross-modal datasets, with potential for further improvements in expert collaboration and adaptive learning.
\clearpage
\section*{Acknowledgments}
This work was supported in part by the National Natural Science Foundation of China (62276120, 62161015, 61966021), and the Yunnan Fundamental Research Projects (202401AS070106, 202301AV070004, 202501AS070123).

{
    \small
    \bibliographystyle{ieeenat_fullname}
    \bibliography{main}
}

\end{document}